\documentclass{article}
\usepackage{spconf,amsmath,graphicx}

\title{CPWC: Contextual Point Wise Convolution for Object Recognition}
\name{Pratik Mazumder$^{1}$\footnotemark[1] \qquad Pravendra Singh$^{1}$\footnotemark[1] \qquad Vinay Namboodiri}

            \address{Department of Computer Science and Engineering, IIT Kanpur, India}
\begin{document}
%
\maketitle
\begin{abstract}
Convolutional layers are a major driving force behind the successes of deep learning. 
Pointwise convolution (PWC) is a $1\times1$ convolutional filter 
that is primarily used for parameter reduction. 
However, the PWC ignores
the spatial information around the points it is processing. This design is by choice, in order to reduce the overall parameters and computations. 
However, we hypothesize that this shortcoming of PWC has a significant impact on the network performance. We propose an alternative design for pointwise convolution, which uses spatial information from the input efficiently. Our design significantly improves the performance of the networks without substantially increasing the number of parameters and computations. 
We experimentally show that our design results in significant improvement in the performance of the network for classification as well as detection.
\end{abstract}
%
%
\section{Introduction and Related Works}
\footnotetext[1]{Equal contribution.}
\label{sec:intro}

Neural networks have helped in successfully dealing with several machine learning and computer vision problems. Convolutional layers have been instrumental in this success. 
As opposed to fully connected layers, the convolutional layers share parameters and capture local spatial context, which enables the network to learn useful features from inputs such as images.
A $1\times1$ convolutional layer (or pointwise convolution) consists of a convolutional filter of size $1\times1$ which works on only one point per channel at a time. 
It was first introduced in Network in Network (NIN) \cite{lin2013network} and was popularised by the Inception network \cite{szegedy2015going}. 
 Pointwise convolutions are widely used in modern architectures to increase or decrease the number of channels in feature maps for computational efficiency.  ResNet/DenseNet \cite{he2016deep,huang2017densely} bottleneck layers use PWC to reduce parameters and computation. 
MobileNetV2 \cite{sandler2018mobilenetv2} use PWCs to increase the number of channels before applying convolutional filters of larger kernel size depthwise and then use them again to decrease the number of channels. 
PWC is used after depthwise and groupwise convolutions in efficient networks such as ShuffleNet \cite{zhang2018shufflenet} and MobileNet \cite{howard2017mobilenets,sandler2018mobilenetv2} to capture channel-wise correlation.

Local spatial context (neighborhood information) is a vital factor behind the successes of convolutional layers. This is especially relevant in cases such as images where spatial neighborhoods contain essential information. 
\textit{However, in pointwise convolution, the local spatial context is totally ignored.}
It is assumed that this has a negligible negative effect.

Our experiments, however, showed that introducing context into pointwise convolutions with a minor increase in parameters/computations can achieve over $1\%$ improvement in the accuracy of the network. 
\textit{Therefore, the performance impact due loss of spatial context is indeed significant}.

We can replace PWCs
with convolutional filters of larger size such as $3\times3$ or $5\times5$, which capture local spatial context. However, doing so will defeat the purpose of pointwise convolution, i.e., to reduce the number of parameters and computations in the network. Therefore, we aim to improve the performance of networks that use PWCs by capturing some spatial context in the PWCs without significantly increasing the parameters/computations.

There has been a lot of works to improve the performance of neural networks. The most common approach is to make a network deeper, i.e., add more layers. This created challenges such as vanishing gradient, which was dealt with by introducing skip connections \cite{he2016deep}. Another approach is to increase the width of network layers, i.e., increasing the number of channels in the layer \cite{zagoruyko2016wide,howard2017mobilenets}. 
Researchers have also used bigger input images/data as they capture more details. But this leads to a significant increase in parameters and computations. Researchers have also pursued other hybrid methods of performance improvement such as \cite{Singh_2019,singh2020hetconv,singh2019accuracy,singh2020cooperative,woo2018cbam}. 


Our aim, however, has been to show that introducing context into pointwise convolution improves the performance of neural networks. \textit{We also propose an efficient way of adding context into pointwise convolution without significantly increasing the number of parameters/computation}. We call our approach the contextual pointwise convolution (CPWC). Our approach can be seen as another way to improve network performance by improving the PWC component of the neural network.
We also show experimentally that our modification to the pointwise convolution results in improved performance that is at par with much deeper/wider models. Our experiments also show that this improvement extends to tasks like object detection in addition to classification.

\begin{figure*}[t]
    \centering
     \includegraphics[width=14cm,height=12cm]{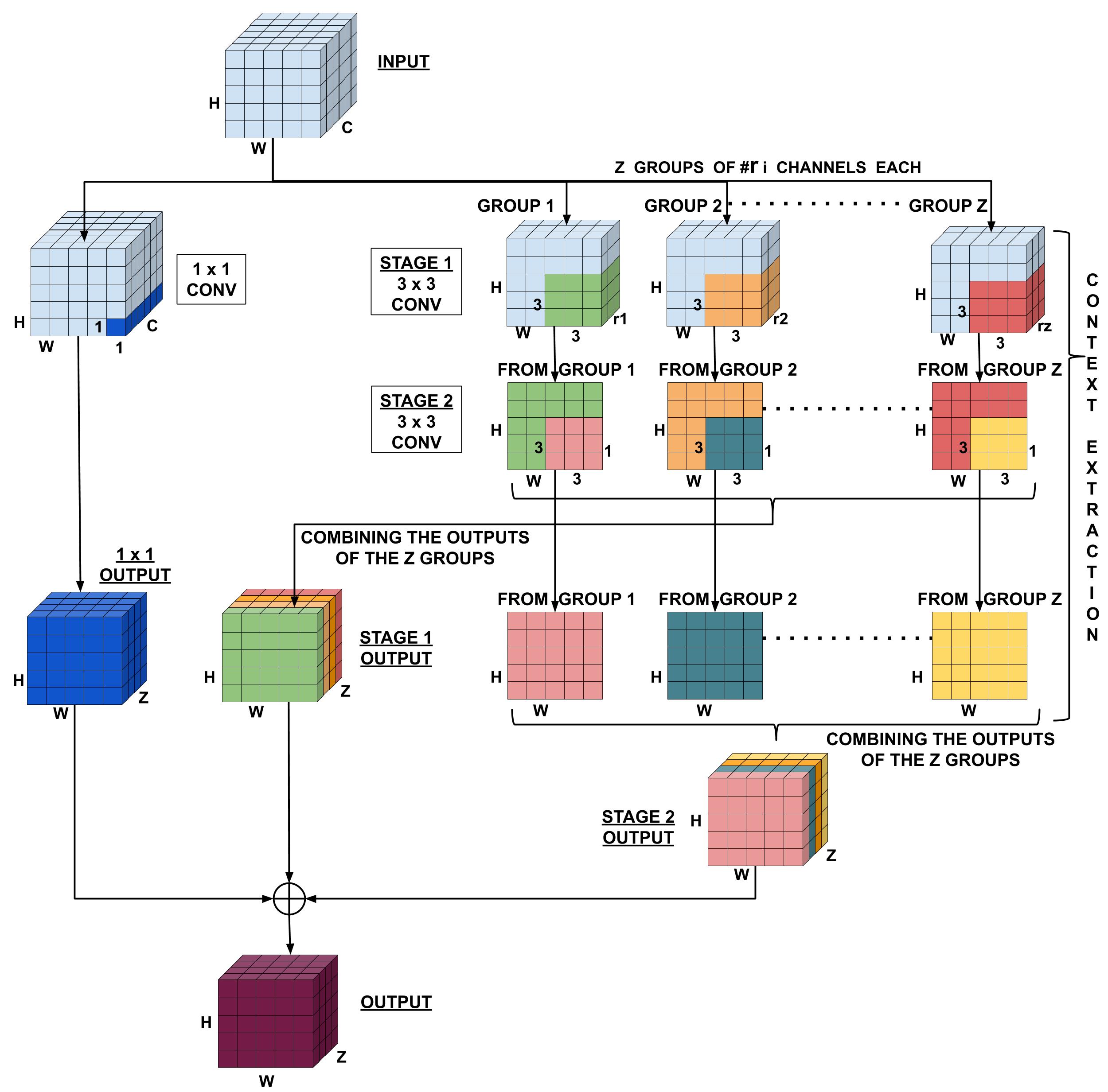}
    \caption{Our proposed approach (CPWC) of context addition to pointwise convolution (best viewed in color). A context extraction unit that extracts spatial context at multiple scales and adds this information to the output of the $1\times1$ convolution.}
    \label{fig:method}
\end{figure*}
\section{Method of Introducing Context}
We propose an efficient approach of adding context to the pointwise convolution. In our approach, the original $1\times1$ convolution is retained and a context extraction unit performs a parallel operation on the input to obtain the local spatial context at multiple scales (stage 1 and stage 2), and this context information gets added to the output of PWC (as shown in Figure \ref{fig:method}). 

As already discussed earlier, the easiest way to get the local context would be to use a convolution filter of larger kernel sizes, such as $3\times3$, $5\times5$, and others. However, using filters of larger kernel size on the input feature maps having a large number of channels will massively increase the network parameters/computations. Therefore, we propose an efficient way of using filters of larger kernel sizes over a few channels in order to get the context and also not to increase the network parameters/computation significantly.

Our objective is to provide the local spatial context information at multiple scales in the form of feature maps that have the same dimensions and number of channels as the output feature maps of the PWC. These feature maps will be added element-wise with the output of the PWC.

Now if there is an improvement in the performance of the network with the modified pointwise convolution block (CPWC) over the original network, we can at least say that our modification has had some positive impact on the network performance. 
We also compare these results with different versions of networks having increased depth or width (and so more parameters than the base network) so that we can check whether this performance improvement is only due to the additional parameters introduced by our modifications. 

Let us take an example from ResNet-50 bottleneck unit, where they reduce the number of channels from 256 to 64. They use PWC to perform this channel reduction (256 to 64), but clearly, this step ignores local spatial context information. We can simply use $3\times3$ convolution to include context information. But this will significantly increase the network parameters/computation because, in order to produce 64 output channels, we require 64 convolutional filters of kernel size $3\times3$. We know that for applying one convolutional filter of kernel size $3\times3$ to an input having 256 channels requires a convolutional filter of actual size $3\times3\times256$. This means that a total of $3\times3\times256\times64$ parameters will be introduced.

To solve the above-mentioned problem, we divide 256 input channels into 64 groups, where each group contains 4 channels. Now we apply a $3\times3$ convolution to each group (stage 1 in Figure~\ref{fig:method}). Therefore, to produce 64 output channels (stage 1 output in Figure~\ref{fig:method}), we need 
64 convolutional filters of actual size $3\times3\times4$. Next, to extract context at a higher scale, we can apply 64 convolutional filters of actual size $3\times3\times1$ (stage 2 in Figure~\ref{fig:method}) on each channel of the stage 1 output to get stage 2 output as shown in Figure~\ref{fig:method}. Finally, we add the stage 1 and stage 2 outputs to the $1\times1$ convolution output in order to incorporate multi-scale context information into the PWC output. 

Please note that in this whole process, stage 1 adds $3\times3\times4\times64$ parameters and stage 2 adds $3\times3\times1\times64$ parameters in addition to the $1\times1$ convolution which adds $1\times1\times256\times64$ parameters. This leads to a total of $19,264$ parameters (stage 1 + stage 2 + PWC). The original PWC adds $16,384$ parameters (only $1\times1$). Whereas if we would have only used 64, $3\times3$ convolutions of actual size $3\times3\times256$ as described earlier, then it would have added $147,456$ parameters.

This idea forms the basis of our approach and also shows how our approach efficiently adds local spatial context information to the PWC output without significantly increasing the number of parameters/computation. As described above, stage 1 and 2 extract contextual information at 2 different scales and provide this information to the PWC output.

\textbf{Selection of Channels in Groups for Stage 1}:
Say we have the input feature maps of total size $W\times H\times C$ where W, H, and C are the width, height, and the number of channels of the input. Let Z be the number of output channels. Therefore, there are Z original $1\times1$ convolutional filters that operate on the input feature maps and finally produce the output feature maps of the total size of $W\times H\times Z$. Stage 1 also has to produce an output of the same size without ignoring any input channel. Therefore $Z$ convolutional filters are required at stage 1. We create $Z$ groups of channels from the input, where the number of channels in the group $i$ is $r_i$, and its size is $W\times H\times r_i$. On every group, one $3\times3$ convolutional filter is applied as shown in  Figure~\ref{fig:method} (stage 1). For example, on group $i$ of size $W\times H\times r_i$, one $3\times3$ convolutional filter of size $3\times 3\times r_i$ is applied, where $r_i$ is the number of channels in group $i$. The number of channels in each group (\textbf{\textit{$r_i$}}) varies according to the number of input and output channels. So we can have 3 cases:

\textbf{Case-1} \textit{Z == C}: When the number of output channels is equal to the number of input channels, then \textbf{\textit{$r_i=1$}} i.e. every group contains only 1 channel exclusively.

\textbf{Case-2} \textit{Z $<$ C}: When the number of output channels is less than the number of input channels, and C is divisible by Z then \textbf{\textit{$r_i=\frac{C}{Z}$}} i.e., every group contain $\frac{C}{Z}$ channels exclusively. If C is not divisible by Z, then let $rm$ be the remainder of the division $\frac{C}{Z}$. In this case the first $rm$ groups will contain $\lfloor\frac{C}{Z}\rfloor + 1$ channels and the rest will contain $\lfloor\frac{C}{Z}\rfloor$ channels each.

\textbf{Case-3} \textit{Z $>$ C}: When the number of output channels is greater than the number of input channels, and  Z is divisible by C then \textbf{\textit{$r_i=1$}}. However, each input channel will be shared by $\frac{Z}{C}$ groups. If Z is not divisible by C, then let $rm$ be the remainder of the division $\frac{Z}{C}$. In this case, each of the first $rm$ channels will be shared by $\lfloor\frac{Z}{C}\rfloor + 1$ groups, and each of the remaining channels will be shared by $\lfloor\frac{Z}{C}\rfloor$ groups. Note that since \textbf{\textit{$r_i=1$}}, each group will contain only 1 input channel but not exclusively.

\textbf{Summary of CPWC}: 
As can be seen in Figure \ref{fig:method}, our method CPWC does not remove the original $1\times1$ convolution but adds a multi-scale spatial context to the PWC output by using a parallel spatial context extraction unit. The spatial context extraction unit consists of Z number of $3\times3$ convolutional filters in stage 1. Each convolutional filter extracts spatial context from the corresponding input group, as described in the section above. Each channel of the output feature maps produced by stage 1 is processed by one $3\times3$ convolutional filter in stage 2. These Z number of $3\times3$ convolutional filters of size $3\times3\times1$, form the stage 2 of context extraction unit which produces stage 2 output as shown in Figure \ref{fig:method}. Once the context extraction is complete, we get 2 output feature maps of total size $W\times H\times Z$, one from each stage. The 2 output feature maps which contain context information are added element-wise to the feature maps produced by the $1\times1$ convolution to get the final output feature maps.

\textbf{Why 2 Stages?} Now, it can be questioned whether stage 1 was enough to obtain the context from the input channels, i.e., whether context from a single scale was enough. However, our ablation studies show that when the context is extracted at two scales (stage 1 and 2), the network performs significantly better than when the context is extracted only at one scale (stage 1). This is also intuitive since by using two $3\times3$ convolutions, we are emulating a $5\times5$ convolution, which, as we know, captures more spatial context information than a $3\times3$ convolution. Further, using two $3\times3$ convolution to emulate a $5\times5$ convolution introduces fewer parameters as compared to using a $5\times5$ convolution \cite{szegedy2015going}. We stop at 2 stages since more stages will add more parameters/computations.
\section{Experiments}
In this section, we report the results of the experiments conducted to find whether the addition of context into $1\times1$ convolution improves the performance of the neural network. We perform this analysis for image classification as well as for object detection. We compare the results of our modified networks to deeper/wider networks to analyze whether the performance improvement due to context addition is solely a result of the increase in parameters. The experiments have all been performed in PyTorch \cite{paszke2017automatic}. In all the architectures, we replace every pointwise convolution with CPWC.
\subsection{Image Classification}
Experiments were conducted for classification on the ImageNet \cite{russakovsky2015imagenet} and CIFAR-100 datasets \cite{krizhevsky2009learning}. 
The training is done on the training set, and the Top-1 accuracy on the validation set is reported. For the ImageNet dataset, experiments 
are performed on the ResNet-50 \cite{he2016deep} architecture and results are also compared with the ResNet-101 architecture. For the CIFAR-100 dataset, experiments are performed on the ResNet-164 \cite{he2016deep}, ResNeXt-29 8$\times$64 \cite{xie2017aggregated} architectures and results are also compared with the ResNet-200 and ResNeXt-29 16$\times$64 architectures. Experiments are also conducted on the efficient architecture design MobileNetV2 \cite{sandler2018mobilenetv2} on the ImageNet and CIFAR-100 datasets.

For experiments on the CIFAR-100 dataset with ResNet-164, we use a learning rate of 0.1, weight decay of 5e-4, momentum of 0.9, and batch size of 128 for 250 epochs and the learning rate is decreased by a factor of 5 for every 50 epochs. For ResNeXt experiments on the CIFAR-100 dataset the standard settings mentioned in the ResNeXt paper \cite{xie2017aggregated} are used.

For experiments on the ImageNet dataset with ResNet-50, we use the same settings as mentioned in the ResNet paper \cite{he2016deep} except for the learning rate which we take as 0.1, the total epochs which we take as 100 and the learning rate is reduced by a factor of 10 for every 30 epochs. The architectures are compared on the basis of accuracy, the number of parameters, and FLOPS. FLoating point OPerations per Second (FLOPS) for a model describes its computational complexity \cite{singh2018stability}. 

\begin{table}[t]
  \begin{center}
    \caption{Classification accuracy ($\%$) on the CIFAR-100 dataset for ResNet-164, ResNeXt-29 and MobileNetV2.}
    \label{tab:cifar100res164}
    \scalebox{.8}{
     \begin{tabular}{|l|c|c|c|} 
      \hline
    \textbf{Models} & \textbf{Params} & \textbf{FLOPS} & \textbf{Acc.}\\
    \hline
    ResNet-164 (Baseline) & $1.74$M & $0.25$G & $77.0$ \\
    \hline
     ResNet-200  & $2.11$M & $0.30$G & $77.5$ \\
    \hline
     ResNet-164 (CPWC w/o Stage 2) & $1.87$M & $0.28$G & $\textbf{77.7}$ \\
    \hline
     \textbf{ResNet-164 CPWC (Ours)} & $1.96$M & $0.30$G & $\textbf{78.4}$ \\
    \hline
   \hline

    \hline
    ResNeXt-29,8$\times$64d (Baseline) & $34.4$M & $5.4$G & $82.23$ \\
    \hline
     ResNeXt-29,16$\times$64d  & $68.1$M & $10.7$G & $82.69$ \\
    \hline
     \textbf{ResNeXt-29,8$\times$64d CPWC (Ours)} & $34.8$M & $5.5$G & $\textbf{82.64}$ \\
    \hline
 
 \hline
 \hline
    MobileNetV2(1.0x) (Baseline) & $2.4$M & $0.30$G & $74.1$ \\
    \hline
    MobileNetV2(1.2x)  & $3.2$M & $0.41$G & $74.3$ \\
    \hline
    
    \textbf{MobileNetV2(1.0x) CPWC (Ours)} & $2.6$M & $0.35$G & $\textbf{75.1}$ \\
    \hline  
    \end{tabular}}
  \end{center}
\end{table}
As can be seen in Table \ref{tab:cifar100res164}, for the CIFAR-100 image classification task, ResNet-164 with CPWC and ResNeXt-29,8$\times$64d with CPWC performs much better than the baselines  ResNet-164 by $1.4\%$ and ResNeXt-29,8$\times$64d respectively. Another interesting point to note is that with CPWC ResNet-164 performs better than ResNet-200, which is a deeper model and ResNeXt-29,8$\times$64d performs similar to  ResNeXt-29,16$\times$64d, which has almost double the number of parameters/FLOPS of the baseline. This shows that introducing context into PWC improves network performance. This also shows that the improvement is not just because of the extra parameters of the network introduced since CPWC perform significantly better than a deeper/wider model.


\begin{table}[t]
  \begin{center}
    \caption{Top-1 Classification accuracy ($\%$) on ImageNet for Resnet-50 and MobileNetV2 (\textbf{re-impl}: re-implementation).}
    \label{tab:imagenetresnet50}
    \scalebox{.8}{
    \addtolength{\tabcolsep}{-4.0pt}
     \begin{tabular}{|l|c|c|c|c|} 
      \hline
    \textbf{Models} & \textbf{Params} & \textbf{FLOPS} & \textbf{Acc.} & \textbf{Acc.}\\
    \textbf{ } & \textbf{ } & \textbf{ } & \textbf{original} & \textbf{re-impl}\\
    \hline
    ResNet-50 (Baseline) & $25.56$M & $4.0$G & $75.3$ & $76.0$ \\
    \hline
     ResNet-101 & $44.55$M & $7.8$G & $76.4$ & $77.2$ \\
    \hline
     ResNet-50 (CPWC w/o Stage 2)  & $25.84$M & $4.2$G  & $-$ &$\textbf{76.6}$ \\
    \hline
     \textbf{ResNet-50 CPWC (Ours)} & $26.05$M & $4.3$G & $-$ & $\textbf{77.2}$ \\
    \hline
    
\hline    
   \hline
    MobileNetV2(1.0x) (Baseline) & $3.4$M & $300$M & $72.0$ & $72.0$ \\
    \hline
    MobileNetV2(1.2x)  & $4.7$M & $465$M & $-$ & $72.4$ \\
    \hline
    \textbf{MobileNetV2(1.0x) CPWC (Ours)} & $3.7$M & $387$M & $-$ & $\textbf{73.0}$ \\
    \hline
    \end{tabular}}
  \end{center}
\end{table}

In Table \ref{tab:imagenetresnet50}, for the ImageNet image classification task, ResNet-50, modified with CPWC, shows a significant improvement in performance over the baseline ResNet-50 with negligible increase in the total network parameters and FLOPS. Our method performs as good as ResNet-101, which is a deeper model and has almost double the number of parameters than the baseline and CPWC.

\textbf{Mobile-optimized networks}: For experiments on CIFAR-100 with MobileNetV2,  we use the same settings as ResNet-164 in the previous section. For experiments on ImageNet, the standard settings mentioned in the MobileNetV2 paper \cite{sandler2018mobilenetv2} are used.
Table \ref{tab:cifar100res164} shows that, for the CIFAR-100 image classification task, MobileNetV2(1.0x) modified with CPWC performs much better than the baseline MobileNetV2(1.0x)  without increasing the total network parameters and FLOPS significantly. Our method performs even better than MobileNetV2(1.2x), which has more parameters and FLOPS than the baseline and our model. 
Similarly, for the ImageNet image classification task (Table \ref{tab:imagenetresnet50}) our approach performs better than the baseline and a wider network MobileNetV2(1.2x) without significantly increasing the parameters/FLOPs. 
\subsection{Object Detection}
\begin{table}[t]
  \begin{center}
    \caption{Object detection mAP ($\%$) on the MS COCO validation set using Faster R-CNN.}
    \label{tab:detection}
    \scalebox{.85}{
    \addtolength{\tabcolsep}{2pt}
     \begin{tabular}{|l|c|c|} 
      \hline
     \textbf{Base Model} & \multicolumn{2}{|c|}{\textbf{AP@IoU =}}\\
     \cline{2-3}
     \textbf{ } & \textbf{0.5:0.95} & \textbf{0.5}\\ 
      \hline
      F-RCNN with ResNet-50 (Baseline) &  $30.3$ & $51.3$ \\
     \hline
     \textbf{F-RCNN with ResNet-50 CPWC (Ours)} &  $\textbf{31.7}$ & $\textbf{52.4}$ \\
     \hline
    \end{tabular}}
  \end{center}
\end{table}
For object detection, we trained the Faster R-CNN \cite{ren2015faster} architecture on the MS-COCO dataset \cite{lin2014microsoft}. 
The ResNet-50 model used in the Faster R-CNN network \cite{ren2015faster} is modified with our CPWC. 
We use a publicly available code for Faster R-CNN with ResNet-50 as a base network \cite{jjfaster2rcnn}. 
In the Faster-RCNN, we use ROI Align and use stride=1 for the last block of the convolutional layer (layer 4) in the base ResNet-50 network.

Table \ref{tab:detection} 
shows a significant improvement in performance in object detection when the context is added to the PWCs used in ResNet-50. This result is significant since context information plays a major role in object detection.

From all the experiments, 
it is clear that adding context into the pointwise convolution has a significant positive impact on the network performance and helps the network perform as well as and even better than heavier models. Therefore, CPWC leads to more efficient performance improvements than by simply increasing the network depth/width.
\section{Ablation Studies}
We validate our proposed approach by performing ablation experiments after removing components from our approach CPWC. We study the effects of removing the stage 2 context extraction unit and the $1\times1$ convolution from CPWC. 

\textbf{Multi-scale context vs Single scale context}:
In our proposed design, we perform context extraction at 2 scales (Stage 1, 2) using $3\times3$ convolutions. As an ablation, we remove Stage 2, thereby effectively performing the context extraction at a single scale only. 
Table \ref{tab:cifar100res164}, \ref{tab:imagenetresnet50} shows a significant reduction in performance when we remove stage 2 from CPWC for both ResNet-164 and ResNet-50 over the CIFAR-100 and ImageNet datasets respectively. Therefore, context extraction at multiple scales  leads to better performance. 

\begin{table}[t]
  \begin{center}
    \caption{Classification accuracy ($\%$) on the CIFAR-100 dataset for ResNet-164 after removing $1\times1$ convolution.}
    \label{tab:ablation1x1}
    \scalebox{.8}{
    \addtolength{\tabcolsep}{8.0pt}
     \begin{tabular}{|l|c|} 
      \hline
    \textbf{Models}  & \textbf{Acc.}\\
    \hline
    ResNet-164 (Baseline)  & $77.0$ \\
    \hline
     ResNet-200  & $77.5$ \\
    \hline
     ResNet-164 (CPWC w/o PWC \& Stage 2 ) & $\textbf{70.8}$ \\
    \hline
    ResNet-164 (CPWC w/o PWC ) & $\textbf{71.6}$ \\
    \hline
     ResNet-164 CPWC (Ours)  & $\textbf{78.4}$ \\
    \hline
   
    \end{tabular}}
  \end{center}
\end{table}

\textbf{Effect of removing the $1\times1$ convolution}:
We removed the $1\times1$ convolution from CPWC, leaving only a context extraction unit that also performs channel reduction. We perform classification on CIFAR-100 dataset with this design.
Table \ref{tab:ablation1x1} shows that the removal of the PWC results in a massive decrease in the performance of CPWC. 
This may be because inside our context extraction unit, the input channels are split into groups and processed as separate groups.
As a result, the channel-wise correlation cannot be fully extracted by such a unit. The PWC performs this role of extracting channel-wise correlation and is hence, vital to our design.
\section{Conclusion}
Through our experiments, we showed that the loss of context due to the use of pointwise convolution ($1\times1$) is significant and should not be ignored. We proposed an efficient design for adding context to the pointwise convolution. We modified the $1\times1$ convolution used in many popular neural networks to incorporate context using our approach. We experimentally showed how our modification resulted in significant improvement in the performance of those networks and brought them at par with deeper/wider models, which used the original $1\times1$ convolution. 
Therefore, our approach delivers a more efficient performance improvement than produced by simply increasing the depth/width of the network. 


\bibliographystyle{IEEEbib}\bibliography{refs}

\begin{thebibliography}{10}

\bibitem{lin2013network}
Min Lin, Qiang Chen, and Shuicheng Yan,
\newblock ``Network in network,''
\newblock {\em arXiv preprint arXiv:1312.4400}, 2013.

\bibitem{szegedy2015going}
Christian Szegedy, Wei Liu, Yangqing Jia, Pierre Sermanet, Scott Reed, Dragomir
  Anguelov, Dumitru Erhan, Vincent Vanhoucke, and Andrew Rabinovich,
\newblock ``Going deeper with convolutions,''
\newblock in {\em Proceedings of the IEEE conference on computer vision and
  pattern recognition}, 2015, pp. 1--9.

\bibitem{he2016deep}
Kaiming He, Xiangyu Zhang, Shaoqing Ren, and Jian Sun,
\newblock ``Deep residual learning for image recognition,''
\newblock in {\em Proceedings of the IEEE conference on computer vision and
  pattern recognition}, 2016, pp. 770--778.

\bibitem{huang2017densely}
Gao Huang, Zhuang Liu, Laurens Van Der~Maaten, and Kilian~Q Weinberger,
\newblock ``Densely connected convolutional networks,''
\newblock in {\em Proceedings of the IEEE conference on computer vision and
  pattern recognition}, 2017, pp. 4700--4708.

\bibitem{sandler2018mobilenetv2}
Mark Sandler, Andrew Howard, Menglong Zhu, Andrey Zhmoginov, and Liang-Chieh
  Chen,
\newblock ``Mobilenetv2: Inverted residuals and linear bottlenecks,''
\newblock in {\em Proceedings of the IEEE Conference on Computer Vision and
  Pattern Recognition}, 2018, pp. 4510--4520.

\bibitem{zhang2018shufflenet}
Xiangyu Zhang, Xinyu Zhou, Mengxiao Lin, and Jian Sun,
\newblock ``Shufflenet: An extremely efficient convolutional neural network for
  mobile devices,''
\newblock in {\em Proceedings of the IEEE Conference on Computer Vision and
  Pattern Recognition}, 2018, pp. 6848--6856.

\bibitem{howard2017mobilenets}
Andrew~G Howard, Menglong Zhu, Bo~Chen, Dmitry Kalenichenko, Weijun Wang,
  Tobias Weyand, Marco Andreetto, and Hartwig Adam,
\newblock ``Mobilenets: Efficient convolutional neural networks for mobile
  vision applications,''
\newblock {\em arXiv preprint arXiv:1704.04861}, 2017.

\bibitem{zagoruyko2016wide}
Sergey Zagoruyko and Nikos Komodakis,
\newblock ``Wide residual networks,''
\newblock {\em arXiv preprint arXiv:1605.07146}, 2016.

\bibitem{Singh_2019}
Pravendra Singh, Vinay~Kumar Verma, Piyush Rai, and Vinay~P. Namboodiri,
\newblock ``Hetconv: Heterogeneous kernel-based convolutions for deep cnns,''
\newblock {\em 2019 IEEE/CVF Conference on Computer Vision and Pattern
  Recognition (CVPR)}, Jun 2019.

\bibitem{singh2020hetconv}
Pravendra Singh, Vinay~Kumar Verma, Piyush Rai, and Vinay~P Namboodiri,
\newblock ``Hetconv: Beyond homogeneous convolution kernels for deep cnns,''
\newblock {\em International Journal of Computer Vision}, pp. 1--21, 2019.

\bibitem{singh2019accuracy}
Pravendra Singh, Pratik Mazumder, and Vinay~P Namboodiri,
\newblock ``Accuracy booster: Performance boosting using feature map
  re-calibration,''
\newblock {\em arXiv preprint arXiv:1903.04407}, 2019.

\bibitem{singh2020cooperative}
Pravendra Singh, Munender Varshney, and Vinay~P Namboodiri,
\newblock ``Cooperative initialization based deep neural network training,''
\newblock {\em arXiv preprint arXiv:2001.01240}, 2020.

\bibitem{woo2018cbam}
Sanghyun Woo, Jongchan Park, Joon-Young Lee, and In~So~Kweon,
\newblock ``Cbam: Convolutional block attention module,''
\newblock in {\em Proceedings of the European Conference on Computer Vision
  (ECCV)}, 2018, pp. 3--19.

\bibitem{paszke2017automatic}
Adam Paszke, Sam Gross, Soumith Chintala, Gregory Chanan, Edward Yang, Zachary
  DeVito, Zeming Lin, Alban Desmaison, Luca Antiga, and Adam Lerer,
\newblock ``Automatic differentiation in pytorch,''
\newblock 2017.

\bibitem{russakovsky2015imagenet}
Olga Russakovsky, Jia Deng, Hao Su, Jonathan Krause, Sanjeev Satheesh, Sean Ma,
  Zhiheng Huang, Andrej Karpathy, Aditya Khosla, Michael Bernstein, et~al.,
\newblock ``Imagenet large scale visual recognition challenge,''
\newblock {\em International Journal of Computer Vision}, vol. 115, no. 3, pp.
  211--252, 2015.

\bibitem{krizhevsky2009learning}
Alex Krizhevsky and Geoffrey Hinton,
\newblock ``Learning multiple layers of features from tiny images,''
\newblock Tech. {R}ep., Citeseer, 2009.

\bibitem{xie2017aggregated}
Saining Xie, Ross Girshick, Piotr Doll{\'a}r, Zhuowen Tu, and Kaiming He,
\newblock ``Aggregated residual transformations for deep neural networks,''
\newblock in {\em Computer Vision and Pattern Recognition (CVPR), 2017 IEEE
  Conference on}. IEEE, 2017, pp. 5987--5995.

\bibitem{singh2018stability}
Pravendra Singh, Vinay Sameer~Raja Kadi, Nikhil Verma, and Vinay~P Namboodiri,
\newblock ``Stability based filter pruning for accelerating deep cnns,''
\newblock {\em WACV}, 2019.

\bibitem{ren2015faster}
Shaoqing Ren, Kaiming He, Ross Girshick, and Jian Sun,
\newblock ``Faster r-cnn: Towards real-time object detection with region
  proposal networks,''
\newblock in {\em Advances in neural information processing systems}, 2015, pp.
  91--99.

\bibitem{lin2014microsoft}
Tsung-Yi Lin, Michael Maire, Serge Belongie, James Hays, Pietro Perona, Deva
  Ramanan, Piotr Doll{\'a}r, and C~Lawrence Zitnick,
\newblock ``Microsoft coco: Common objects in context,''
\newblock in {\em European conference on computer vision}. Springer, 2014, pp.
  740--755.

\bibitem{jjfaster2rcnn}
Jianwei Yang, Jiasen Lu, Dhruv Batra, and Devi Parikh,
\newblock ``A faster pytorch implementation of faster r-cnn,''
\newblock {\em https://github.com/jwyang/faster-rcnn.pytorch}, 2017.

\end{thebibliography}

\end{document}